\documentclass{template/llncs}

\usepackage{template/llncsdoc}

\usepackage{graphicx}
\usepackage{amssymb,amsmath,bm}
\usepackage{textcomp}
\usepackage{url}
\usepackage[nolist]{acronym}

\begin{acronym}
    \acro{SVM}{Support Vector Machine}
    \acro{RF}{Random Forest}
    \acro{ASR}{Automatic Speech Recognition}
    \acro{CMU}{Carnegie Mellon University}
    \acro{WER}{Word Error Rate}
    \acro{SMO}{Sequential Minimal Optimization}
    \acro{FCT}{Funda\c{c}\~{a}o para a Ci\^{e}ncia e a Tecnologia}
    \acro{KTH}{KTH Royal Institute of Technology}
\end{acronym}

\begin{document}

\thispagestyle{empty}

\title{Assessing User Expertise in Spoken Dialog System Interactions}
\author{Eug\'{e}nio Ribeiro\inst{1,2} \and Fernando Batista\inst{1,3} \and Isabel Trancoso\inst{1,2}
\and Jos\'{e} Lopes\inst{4} \and Ricardo Ribeiro\inst{1,3} \and David Martins de Matos\inst{1,2}}
\institute{L$^2$F {--} Spoken Language Systems Laboratory {--} INESC-ID Lisboa
\and
Instituto Superior T\'{e}cnico, Universidade de Lisboa, Portugal
\and
ISCTE-IUL {--} Instituto Universit\'{a}rio de Lisboa, Portugal
\and
KTH Speech, Music, and Hearing, Stockholm, Sweden
\email{eugenio.ribeiro@l2f.inesc-id.pt}}

\maketitle

%%%
%
%   ABSTRACT
%

\begin{abstract}

Identifying the level of expertise of its users is important for a system since it can lead to a better interaction through adaptation techniques. Furthermore, this information can be used in offline processes of root cause analysis. However, not much effort has been put into automatically identifying the level of expertise of an user, especially in dialog-based interactions. In this paper we present an approach based on a specific set of task related features. Based on the distribution of the features among the two classes {--} Novice and Expert {--} we used Random Forests as a classification approach. Furthermore, we used a Support Vector Machine classifier, in order to perform a result comparison. By applying these approaches on data from a real system, Let's Go, we obtained preliminary results that we consider positive, given the difficulty of the task and the lack of competing approaches for comparison.

\keywords{user expertise $\cdot$ Let's Go $\cdot$ SVM $\cdot$ Random Forest}

\end{abstract}

%
%   END OF ABSTRACT
%
%%%

%%%
%
%   INTRODUCTION
%

\section{Introduction}
\label{sec:introduction}

The users of a dialog system have different levels of expertise, that is, knowledge of the system's capabilities and experience using it. Thus, identifying the level of expertise of a user is important for a dialog system, since it provides cues for adaptation which can improve dialog flow and the overall user satisfaction. For instance, by identifying a novice user, the system may provide help on the first signs of struggle and adapt its prompts to provide further information. Also, user expertise information can be used to adapt the system's parameters, such as \ac{ASR} timeout values, reducing the number of misinterpretations and interruptions. Furthermore, it can be used in offline processes to identify problems caused by lack of expertise, which is important for the development of better dialog systems.   

In this article we present an analysis of different features and how they can be used to identify the level of expertise of a user on Let's Go~\cite{Raux2006} data. The remaining sections are structured as follows: Section~\ref{sec:related} presents related work on user expertise with dialog systems. Section~\ref{sec:features} lists relevant feature classes for this task. Section~\ref{sec:setup} describes the datasets, the specific features extracted, and the classification approaches. Results are presented and discussed in Section~\ref{sec:results}, and, finally, Section~\ref{sec:conclusions} states the achieved conclusions and suggests paths for future work.

%
%   END OF SECTION
%
%%%

%%%
%
%   RELATED WORK
%

\section{Related Work}
\label{sec:related}

A system that behaves the same way for all users, independently of their expertise, may not provide a truly usable interface for any of them. By knowing the level of expertise of its users, a system could improve the quality of the interaction through adaptation techniques based on that level~\cite{Nielsen1993}. However, not much effort has been put into identifying the level of expertise of a user, especially in dialog-based interactions.

Hjalmarsson~\cite{Hjalmarsson2005b} analyzed dialog dynamics and discussed the utility of creating adaptive spoken dialog systems and individual user models. She suggests that such models can be created using both rule-based and statistical approaches~\cite{Hjalmarsson2005a}. Given the correct set of rules, rule-based models have good performance on specific tasks. However, they must be handcrafted from the intuition of the designer or experts, which is a time-consuming process. Thus, when annotated data is available, statistical models are a better option. The author suggests Bayesian Networks, Reinforcement Learning, and Decision Trees as promising approaches for the task.

Hassel and Hagen~\cite{Hassel2006} developed an automotive dialog system that adapts to its users' expertise. However, it does not attempt to identify each user's expertise, but rather assumes that every user is a novice and then adapts over time. The adaptation is task-based and controlled by the number of successful attempts and the time since the last execution of that task.

Jokinen and Kanto~\cite{Jokinen2004} used user expertise modelling to enable the adaptation of a speech-based e-mail system. They distinguish three levels of expertise {--} Novice, Competent, and Expert {--}, a subset of the five proposed by Dreyfus and Dreyfus~\cite{Dreyfus1986} in their studies about the behaviour of expert systems. The developed models have two components {--} online and offline. The first tracks the current user session and provides cues for system adaptation, accordingly. The latter is based on statistical event distributions created from all sessions of a user and serves as a starting point for the next session. In terms of features, a small set is used, consisting on the number of timeouts, interruptions, and help requests, as well as the number of times a given task was performed, or a specific system dialog act was invoked.

%
%   END OF SECTION
%
%%%

%%%
%
%   FEATURES
%

\section{Relevant Features}
\label{sec:features}

Since expertise depends on the task being performed, it cannot be identified by large sets of generic acoustic features such as the ones extracted by openSMILE~\cite{Eyben2013}. Thus, a small set of task oriented features must be devised. These features can be clustered into different categories, according to their origin and what aspects they intend to capture. In the following sections we describe each of these categories.

\subsection{Interruptions}
\label{ssec:interruptions}

Expert users may interrupt the system when they are aware of the complete dialog flow. However, this is not a good approach when system utterances are confirmation prompts which only include the information to be confirmed in the end. In this case, interruptions usually signal a novice user. Furthermore, cases when the system interrupts the user can also reveal a novice user who uses long sentences or pauses that exceed the system's waiting times.

\subsection{Delays}
\label{ssec:delays}

Negative delays between the system and user utterances mean that an interruption occurred, which has the previously described implications. On the other hand, long delays suggest that the user is still processing the system's utterance and is unsure about what to say and, thus, may reveal inexperience.

\subsection{Durations}
\label{ssec:durations}

Long durations are typically more discriminative than short ones and may suggest inexperience. For instance, a long call usually means that something went wrong during the dialog. Long utterances also suggest inexperience, as they are more prone to recognition errors and interruptions by the system.

\subsection{Speech Rate}
\label{ssec:sr}

Speech rate is also a possible indicator of the level of expertise of a user since both high and low speech rates may lead to communication problems. While high speech rates lead to higher error rates in recognition, low speech rates are related to paused speeches, which are more prone to be interrupted by the system. Thus, expert users usually keep a balanced speech rate.

\subsection{Help Requests}
\label{ssec:help}

When a user is new to a system and unsure how it works, he or she typically asks for help, revealing inexperience. This is especially clear in cases when the system offers help and the user immediately accepts it. Unfortunately, some systems do not provide help functionality or the user is not aware it exists.

%
%   END OF SECTION
%
%%%

%%%
%
%   EXPERIMENTAL SETUP
%

\section{Experimental Setup}
\label{sec:setup}

This section describes our experimental setup, starting with the used datasets. Next, the used features and their distribution in the training dataset are thoroughly presented. After that, the used classification and evaluation approaches are described.

%%%
%
%   DATASETS
%

\subsection{Datasets}
\label{ssec:datasets}

We explored user expertise on data extracted from interactions with the Let's Go Bus Information System~\cite{Raux2006}, which provides information about bus schedules, through spoken telephonic interaction with a dialog system. This system has been running for many years and has experienced changes over time. Thus, the characteristics of the data differ according to when it was obtained.

In our experiments we used the LEGO~\cite{Schmitt2012} corpus. This corpus is a subset of 347 Let's Go calls during the year of 2006. The corpus contains annotations relevant for the user level of expertise identification task, such as barge-in information and durations. In terms of the level of expertise, the original corpus is not annotated. Thus, we annotated each call with that information using two labels {--} Expert and Novice. Out of the 347 calls, 80 users were labeled as Expert and 235 as Novice. The remaining calls were impossible to annotate since the user did not interact with the system. We used this dataset for analyzing the distribution of different features and as training data for the classification task.

In addition to the 2006 data, we also looked into a small set of 80 calls from 2014 data of the Let's Go corpus. This set was annotated for expertise at KTH using the same two labels {--} Expert and Novice~\cite{Lopes2016}. The audio files for all but one of calls are accompanied by the logs of the system, which provide important information for feature extraction. Of the 79 calls, 42 users were labeled as Expert and 37 as Novice. The reported Cohen's Kappa for the annotator agreement was 0.73, which is usually considered good. However, we also annotated the 79 calls, labeling 31 users as Expert and 48 as Novice, obtaining an agreement of 0.43 with the original annotation, which is usually considered moderate. We used the 56 calls with agreement to assess the generalization capabilities of our classifiers.

%
%   END OF SUBSECTION
%
%%%

%%%
%
%   FEATURES
%

\subsection{Features}
\label{ssec:features}

In Section~\ref{sec:features} we defined a set of feature classes that are relevant for identifying the level of expertise of a user. In this section we describe the specific features that we were able to extract from the datasets. Furthermore, for the training dataset, LEGO, we perform an analysis of the distributions of the features among the data, in order to perform a comparison with the previously defined expectations. Table~\ref{tab:featdistr} presents these distributions.

\begin{table}[htbp]
\caption{Feature distribution among the LEGO dataset in terms of average ($\mu$), median(\~{x}), and standard deviation($\sigma$).}
\label{tab:featdistr}
\centering
\begin{tabular}{|l|rrr|rrr|}
\cline{2-7}
\multicolumn{1}{c}{} & \multicolumn{3}{|c|}{\textbf{Novice}} & \multicolumn{3}{|c|}{\textbf{Expert}} \tabularnewline
\cline{1-1}
\textbf{Feature} & \multicolumn{1}{|c}{$\mu$} & \multicolumn{1}{c}{\~{x}} & \multicolumn{1}{c|}{$\sigma$} & \multicolumn{1}{|c}{$\mu$} & \multicolumn{1}{c}{\~{x}} & \multicolumn{1}{c|}{$\sigma$} \tabularnewline
%\textbf{Feature} & \multicolumn{1}{|c}{Average} & \multicolumn{1}{c}{Median} & \multicolumn{1}{c|}{StDev} & \multicolumn{1}{|c}{Average} & \multicolumn{1}{c}{Median} & \multicolumn{1}{c|}{StDev} \tabularnewline
\hline
\multicolumn{7}{|l|}{\textbf{Interruptions}} \tabularnewline
\hline
\# Barge-ins                        &  5.06 &  3.00 &     6.79 &  2.75 &  2.00 &     3.15 \tabularnewline
Barge-in Rate                        &  16.2 &  15.4 &      9.9 &  10.3 &   9.5 &      6.9 \tabularnewline
\hline
\multicolumn{7}{|l|}{\textbf{Delays (s)}} \tabularnewline
\hline
1\textsuperscript{st} Turn          &  1.52 &  1.28 &     3.00 &  1.32 &  1.21 &     2.81 \tabularnewline
1\textsuperscript{st} Turn (Positive)      &  2.82 &  2.18 &     2.79 &  1.90 &  1.49 &     2.72 \tabularnewline
\hline
\multicolumn{7}{|l|}{\textbf{Durations (s)}} \tabularnewline
\hline
Utterance                           &  1.81 &  1.44 &     3.14 &  1.19 &  1.20 &     0.43 \tabularnewline
Call                                &   123 &   104 &       95 &   102 &    76 &       78 \tabularnewline
1\textsuperscript{st} Turn          &  1.81 &  1.19 &     2.02 &  1.72 &  1.39 &     1.66 \tabularnewline
\# Exchanges                        &  28.0 &  23.0 &     23.4 &  23.8 &  20.0 &     13.8 \tabularnewline
\hline
\multicolumn{7}{|l|}{\textbf{Speech Rate (phones/s)}} \tabularnewline
\hline
Global                              &  13.7 &  14.2 &      3.3 &  14.8 &  14.9 &      1.9 \tabularnewline
1\textsuperscript{st} Turn          &  14.3 &  14.5 &      4.1 &  14.8 &  14.5 &      2.8 \tabularnewline
\hline
\multicolumn{7}{|l|}{\textbf{Help Requests}} \tabularnewline
\hline
\# Requests                         &  0.27 &  0.00 &     0.55 &  0.00 &  0.00 &     0.00 \tabularnewline
\hline
\end{tabular}
\end{table}

\subsubsection{Interruptions}
\label{sssec:interruptions}

The LEGO corpus is annotated with user barge-in information. Thus, we were able to extract the number of user barge-ins per dialog. Table~\ref{tab:featdistr} shows that novice users are more prone to interrupt the system, with an average of 5 barge-ins per dialog. This was expected, since most of the system utterances in the corpus are of the kind that state information to be confirmed in the final words and, thus, should not be interrupted. However, these statistics did not take the length of the dialog into account. Thus, we calculated the user barge-in rate as the percentage of exchanges containing user barge-ins. The previous results were confirmed, as, on average, novice users barged-in on 16\% of the exchanges, while experts only barged-in on 10\% of the exchanges. The median values of 15\% and 10\% for novice and expert users, respectively, also support the hypothesis. Furthermore, only novice users have barge-in rates over 30\%. Information extracted from the first turn is not as discriminative, as around 60\% of the users barged-in on the first turn, independently of the class. However, the first system utterance is a fixed introduction, which encourages expert users to barge-in, in order to skip to important parts of the dialog.

On Let's Go 2014, the barge-in information was extracted from the interaction logs.

\subsubsection{Delays}
\label{sssec:delays}

LEGO annotations include the duration of each user utterance, as well as the time when each exchange starts. However, each exchange starts with the system utterance, for which the duration is not annotated. Thus, we were not able to obtain delay information for most exchanges. The only exception was the first exchange, since the system utterance is fixed and, thus, so is its duration {--} 10.25 seconds. In this case, we calculated the delay as the difference between the time when the user utterance starts and the time when the system utterance ends. As expected, the results presented in Table~\ref{tab:featdistr} suggest that novice users take longer to answer than expert users. Furthermore, when only positive delay values are taken into account, the discrepancy between the two classes is even more evident.

On the 2014 data, we used a similar approach to obtain the first turn delay. In this case, the duration of the first system utterance is 13.29 seconds. The remaining information required to calculate the delay was obtained from the interaction logs.

\subsubsection{Durations}
\label{sssec:durations}

The LEGO corpus is annotated in terms of duration of user utterances, as well of the whole call. However, a few of the utterances are wrongly annotated. Nonetheless, we were able to compute the average user utterance duration per dialog. As expected, novice users tend to use longer utterances and are much less consistent than expert users. There are no expert users with average utterance durations over 3 seconds. In terms of the whole call, the same conclusions can be drawn, both in terms of time duration and number of exchanges, as novice users have higher values for all the measures. While most calls by expert users last less than 2 minutes, calls by novice users have a wider distribution. As for the duration of the first utterance, on average, novice users still use longer utterances. However, that is not true in terms of median value. Nonetheless, standard deviation for novice users is higher than the average value, which suggests that novice users adopt unpredictable behaviors.

We obtained duration information from 2014 data directly from the audio files, using SoX~\cite{sox}. %~\footnote{\url{http://sox.sourceforge.net/}}.

\subsubsection{Speech Rate}
\label{sssec:sr}

We extracted the speech rate in phones per second from each user utterance of the LEGO corpus and used those values to calculate the average speech rate for each dialog. The phones for each utterance were obtained using the neural networks included in the AUDIMUS~\cite{Meinedo2008} \ac{ASR} system. Table~\ref{tab:featdistr} shows similar average and median values for both classes, around 15 phones per second. However, expert users are more steady, which leaves the tails of the distribution for novice users only. Looking only at the first user utterance, average and median values are even closer for both classes. Nonetheless, the tails of the distribution are still reserved for novice users only, although the expert users are slightly less steady. The same extraction procedure was applied on 2014 data.

\subsubsection{Help Requests}
\label{sssec:help}

From the existing information, we were able to extract the number of help requests detected by the system during each LEGO dialog. As expected, only novice users asked for help, with an average of 0.27 help requests per dialog. 23\% of novice users asked for help at least once and up to 3 times. Furthermore, 17\% of the novice users asked for help on the first turn.

On the 2014 data, we obtained the number of help requests from the dialog transcriptions, by looking for the help keyword or the zero key on user utterances.

%
%   END OF SUBSECTION
%
%%%

%%%
%
%   APPROACHES
%

\subsection{Classification}
\label{ssec:approaches}

Distinguishing between novice and expert users is a binary classification task. From the multiple classification approaches that could be used, we opted \acp{SVM}\cite{Cortes1995}, since it is a widely used approach and typically produces acceptable results, and \ac{RF}~\cite{Breiman2001}, an approach based on decision trees, which are indicated for this task, given the distribution of our features among the two classes.

To train our \acp{SVM}, we took advantage of the \ac{SMO} algorithm~\cite{Platt1998} implementation provided by the Weka Toolkit~\cite{Hall2009}. We used the linear kernel and kept the C parameter with its default 1.0 value.

We opted for an \ac{RF} approach due to its improved performance when compared to a classic decision tree algorithm. We also used the implementation provided by the Weka Toolkit to train our \acp{RF}. We used 1000 as the number of generated trees, since it provided a good trade-off between training time and classification accuracy. 

%
%   END OF SUBSECTION
%
%%%

%%%
%
%   EVALUATION
%

\subsection{Evaluation}
\label{ssec:evaluation}

Since there is no standard partition of the LEGO corpus into training and testing sets, we obtained results using 10-fold cross-validation. Furthermore, we used the data from 2014 to assess the generalization capabilities of our classifiers.

In terms of measures, we use Accuracy and the Kappa Statistic since they are the most indicated measures to evaluate performance and relevance on this task. Accuracy is given by the ratio between the number of correct predictions and the total number of predictions. The Kappa Statistic gives the weighted agreement between the predictions of the classifier and the gold standard, in relation to those of a chance classifier.

%
%   END OF SUBSECTION
%
%%%

%
%   END OF SECTION
%
%%%

%%%
%
%   RESULTS
%

\section{Results}
\label{sec:results}

Since the LEGO dataset is highly unbalanced, we balanced it using the Spread Subsample filter provided by the Weka Toolkit. Still, we performed experiments on both the balanced and unbalanced data. Table~\ref{tab:results} presents the results obtained using each set of features independently, as well as different combinations. The \textbf{First Turn} set combines the features extracted from the first turn only, while the \textbf{Global} set combines the features extracted from the whole dialog. The \textbf{All} set combines the two previous sets. The \textbf{Selected} set is obtained by applying the Best First feature selection algorithm, provided by the Weka Toolkit, to the \textbf{All} set.

\begin{table}[htbp]
\caption{Results on the unbalanced (Chance = 0.741) and balanced (Chance = 0.500) versions of the LEGO dataset}
\label{tab:results}
\centering
\begin{tabular}{|l|cc||cc|cc|}
\cline{2-7}
\multicolumn{1}{c|}{} & \multicolumn{2}{c||}{\textbf{Unbalanced}} & \multicolumn{4}{c|}{\textbf{Balanced}} \tabularnewline
\cline{2-7}
\multicolumn{1}{c|}{} & \multicolumn{2}{c||}{\textbf{Random Forest}} & \multicolumn{2}{c|}{\textbf{SVM}} & \multicolumn{2}{c|}{\textbf{Random Forest}} \tabularnewline
\cline{1-1}
\textbf{Feature Set} & Accuracy & $\kappa$ & Accuracy & $\kappa$ & Accuracy & $\kappa$ \tabularnewline
\hline
Interruptions   & 0.702 & 0.140 & 0.600 & 0.200 & 0.613 & 0.225 \tabularnewline
Delays          & 0.693 & 0.168 & 0.494 & -0.013 & 0.519 & 0.038 \tabularnewline
Durations       & 0.790 & \textbf{0.403} & 0.594 & 0.188 & 0.744 & 0.488 \tabularnewline
Speech Rate     & 0.686 & 0.037 & 0.513 & 0.025 & 0.525 & 0.050 \tabularnewline
Help Requests   & 0.741 & 0.000 & 0.631 & 0.263 & 0.600 & 0.200 \tabularnewline
\hline
\textbf{First Turn}     & 0.767 & 0.321 & 0.594 & 0.188 & 0.713 & 0.425 \tabularnewline
\textbf{Global}         & 0.783 & 0.377 & 0.681 & 0.363 & 0.769 & 0.538 \tabularnewline
\textbf{All}            & 0.793 & 0.385 & \textbf{0.706} & \textbf{0.413} & \textbf{0.794} & \textbf{0.588} \tabularnewline
\textbf{Selected}       & \textbf{0.796} & \textbf{0.403} & \textbf{0.706} & \textbf{0.413} & 0.781 & 0.563 \tabularnewline
\hline
\end{tabular}
\end{table}

The \acp{SVM} classification approach performed poorly on the unbalanced dataset, never surpassing a chance classifier. However, the \ac{RF} approach achieved 80\% accuracy using the \textbf{Selected} feature set, which represents an improvement of 6 percentage points. Given the difficulty and subjectivity of the task, the Kappa coefficient of 0.40 should not be disregarded.

On the balanced dataset, both the \ac{SVM} and \ac{RF} approaches were able to surpass the chance classifier. Still, similarly to to what happened on the unbalanced dataset, the \ac{RF} approach performed better. Using all the available features, it achieved 79\% accuracy, which represents an improvement of 8 percentage points over the \ac{SVM} counterpart and 29 percentage points over the chance classifier. The Kappa coefficient of 0.59 is 50\% higher than the one obtained for the unbalanced dataset, in spite of facing the same concerns. In this version of the dataset, feature selection did not improve the results.

The \textbf{First Turn} feature set is the most relevant for expertise level identification in real time. Using this set, an accuracy of 77\% was achieved on the unbalanced dataset, which represents an improvement of 3 percentage points over the chance classifier. On the balanced dataset, the \ac{RF} approach was able to improve the results of a chance classifier by 21 percentage points and achieve a Kappa coefficient of 0.42. However, the \ac{SVM} classifier performed poorly. Overall, this means that it is not easy to identify the level of expertise of a user based solely on the first turn of the dialog. Still, a preliminary classification can be obtained to start guiding the system towards user adaptation, and improved as the dialog flows.

In terms of the individual feature sets, duration related features are the most important for the \ac{RF} approach on both versions of the dataset. On the balanced dataset, interruption and help related features also provide important information. For the \ac{SVM} approach, the important features remain the same but the order of importance is inverted.

Table~\ref{tab:results2014} presents the results obtained on Let's Go 2014 data by the classifiers trained on the balanced LEGO corpus. We do not show the rows related to feature categories that did not provide relevant results. We can see that, in this case, the \ac{SVM} approach surpassed the \ac{RF} one, achieving 66\% accuracy and a Kappa coefficient of 0.33, using the \textbf{Selected} feature set. This represents an improvement of 11 percentage points over the chance classifier. As for the \ac{RF} approach, although its accuracy using the \textbf{Selected} feature set is just two percentage points below the \ac{SVM} approach, its Kappa coefficient of 0.22 is much lower and is surpassed, although only slightly, by the 0.23 obtained by using only help related features. Overall, this means that the \ac{RF} classifiers, which performed better on the LEGO corpus, have less generalization capabilities than the \ac{SVM} ones. This explains the negative results obtained by the \ac{RF} classifier using the \textbf{Global} feature set, as the differences between both datasets are more noticeable when looking at the dialogs as a whole than when just looking at first turns.

\begin{table}[htbp]
\caption{Results on Let's Go 2014 data (Chance = 0.554)}
\label{tab:results2014}
\centering
\begin{tabular}{|l|cc|cc|}
\cline{2-5}
\multicolumn{1}{c|}{} & \multicolumn{2}{c|}{\textbf{SVM}} & \multicolumn{2}{c|}{\textbf{Random Forest}} \tabularnewline
\cline{1-1}
\textbf{Feature Set} & Accuracy & $\kappa$ & Accuracy & $\kappa$ \tabularnewline
\hline
%Interruptions   & 0.554 & 0.093 & 0.518 & 0.013 \tabularnewline
%Delays          & 0.446 & 0.000 & 0.589 & 0.191 \tabularnewline
%Durations       & 0.464 & 0.029 & 0.607 & 0.139 \tabularnewline
%Speech Rate     & 0.464 & -0.110 & 0.518 & 0.021 \tabularnewline
Help Requests   & 0.607 & 0.268 & 0.589 & \textbf{0.232} \tabularnewline
\hline
\textbf{First Turn}     & 0.571 & 0.207 & 0.538 & 0.082 \tabularnewline
\textbf{Global}         & 0.589 & 0.153 & 0.538 & -0.018 \tabularnewline
\textbf{All}            & 0.643 & 0.283 & 0.589 & 0.103 \tabularnewline
\textbf{Selected}       & \textbf{0.661} & \textbf{0.327} & \textbf{0.643} & 0.217 \tabularnewline
\hline
\end{tabular}
\end{table}

%
%   END OF SECTION
%
%%%

%%%
%
%   CONCLUSIONS
%

\section{Conclusions}
\label{sec:conclusions}

In this article we presented an approach for automatically distinguishing novice and expert users based on a specific set of task related features. Given the distributions of the features, a classification approach based on decision trees was indicated. This was confirmed when the \ac{RF} approach outperformed the widely used \acp{SVM} on both versions of the LEGO corpus.

Since this is a relatively unexplored task and the dataset was not previously annotated for expertise, we cannot compare our results with other work. Nonetheless, we believe that the obtained results are positive, since our approach focused on identifying the level of expertise from a single session, without previous information about the user, which is a difficult task.

Furthermore, we were also able to obtain relevant results using features extracted only from the first turn of each dialog. This is important for a fast adaptation of the system to the user's level of expertise, as it provides a preliminary classification of that level, which can be improved as the dialog flows, through the accumulation of the results of all turns.

On the downside, the results obtained on the data from Let's Go 2014 were not as satisfactory, with the \ac{RF} classifiers revealing less generalization capabilities than the \ac{SVM} ones.

In terms of future work, we believe that it would be important to obtain more annotated data, in order to train more reliable classifiers, with improved generalization capabilities.

%
%   END OF SECTION
%
%%%

%%%
%
%   ACKNOWLEDGEMENTS
%

\section*{Acknowledgements}
\label{sec:acknowledgements}

This work was supported by national funds through Funda\c{c}\~{a}o para a Ci\^{e}ncia e a Tecnologia (FCT) with reference UID/CEC/50021/2013, by Universidade de Lisboa, and by the EC H2020 project RAGE under grant agreement No 644187.

%
%   END OF SECTION
%
%%%

\bibliographystyle{template/splncs03}

\bibliography{references}

\end{document}